\documentclass[11pt, a4paper, logo, copyright, nonumbering]{md_from_ds}
\usepackage[numbers, sort&compress]{natbib}

\usepackage{dblfloatfix}
\usepackage{ulem}
\usepackage{dramatist}
\usepackage{xspace}
\usepackage{pifont}
\usepackage{multirow}
\usepackage{tcolorbox}
\usepackage{xltabular}
\usepackage{longtable}
\usepackage{pdfpages}
\definecolor{linkcolor}{rgb}{0.956,0.298,0.235}
\definecolor{citecolor}{HTML}{1976D2}
\usepackage{tabularx}
\usepackage{hyperref}
\usepackage{graphicx}
\usepackage{subfigure}
\usepackage{amsmath}
\usepackage{amssymb}
\usepackage{array}
\usepackage{booktabs}
\usepackage{multirow}
\usepackage{arydshln}
\usepackage[utf8]{inputenc}
\hypersetup{
    colorlinks=true,
    linkcolor=citecolor,
    filecolor=magenta,
    urlcolor=cyan,
    citecolor=citecolor,
}

\usepackage{amsfonts}
\usepackage{amsmath}
\usepackage{amssymb}
\usepackage{lineno}

\usepackage[bottom]{footmisc}

\usepackage{CJKutf8}
\usepackage{subfigure}
\usepackage{setspace}

\usepackage{xspace}

\usepackage{tcolorbox}
\usepackage{colortbl}
\usepackage{geometry}
\usepackage{caption}

\makeatletter
\def\@BTrule[#1]{
  \ifx\longtable\undefined
    \let\@BTswitch\@BTnormal
  \else\ifx\hline\LT@hline
    \nobreak
    \let\@BTswitch\@BLTrule
  \else
     \let\@BTswitch\@BTnormal
  \fi\fi
  \global\@thisrulewidth=#1\relax
  \ifnum\@thisruleclass=\tw@\vskip\@aboverulesep\else
  \ifnum\@lastruleclass=\z@\vskip\@aboverulesep\else
  \ifnum\@lastruleclass=\@ne\vskip\doublerulesep\fi\fi\fi
  \@BTswitch}
\makeatother

\addto\extrasenglish{

}

\newlength\savewidth

\reportnumber{001}

\renewcommand{\today}{}

\title{\centering  EMMA: Efficient Multimodal Understanding, Generation, and Editing with a Unified Architecture}

\author[*]{
\small

Xin He$^*$, Longhui Wei$^{*\dagger\ddagger}$, Jianbo Ouyang$^*$, Minghui Liao, Lingxi Xie, Qi Tian$^\ddagger$

\small
$*$ Equal Contribution \quad $\dagger$ Project Lead  \quad $\ddagger$ Corresponding Author 

\small
weilh2568@gmail.com, tian.qi1@huawei.com

\small
Huawei Inc. \\

\small
Project Page: \url{https://emma-umm.github.io/emma/}
\small
}

\begin{abstract}\fontsize{11pt}{12pt}
    We propose \textbf{EMMA}, an efficient and unified architecture for multimodal understanding, generation and editing. Specifically, EMMA primarily consists of 1) An efficient autoencoder with a 32x compression ratio, which significantly reduces the number of tokens required for generation. This also ensures the training balance between understanding and generation tasks by applying the same compression ratio to images. 2) Channel-wise concatenation instead of token-wise concatenation among visual understanding and generation tokens, which further reduces the visual tokens in unified architectures. 3) A shared-and-decoupled network that enables mutual improvements across tasks while meeting the task-specific modeling requirements. 4) A mixture-of-experts mechanism adopted for visual understanding encoder, which substantially improves perceptual capabilities with a few parameters increase. {Extensive experiments have shown that EMMA-4B can significantly outperform state-of-the-art unified multimodal approaches (\textit{e.g.}, BAGEL-7B) in both efficiency and performance, while also achieving competitive results compared to recent multimodal understanding and generation experts (\textit{e.g.}, Qwen3-VL and Qwen-Image)}. We believe that EMMA lays a solid foundation for the future development of unified multimodal architectures.

\end{abstract}

\begin{document}
\begin{CJK*}{UTF8}{gbsn}

\maketitle

\section{Introduction}

Recently, unified multimodal architectures have garnered unprecedented attention. A growing number of researchers have recognized that pushing the boundaries and limitations of unified multimodal foundation models is crucial to the progress of multimodal research, embodied artificial intelligence, and even artificial general intelligence~\cite{wang2024emu3,emu35,wu2024janus,li2025dual,deng2025BAGEL,he2024incorporating,wei2022mvp}. Consequently, unified multimodal models have seen rapid development. 

Currently, unified multimodal approaches~\cite{wu2024janus,chen2025januspro,lin2025uniworld,deng2025BAGEL,xie2024showo,chen2025blip3o,wang2024emu3} can be broadly categorized into three main directions: 1) \textbf{Unification of architecture formats}: These approaches (such as BAGEL~\cite{deng2025BAGEL} and JanusFlow~\cite{ma2025janusflow}) facilitate the deep interaction and integration between understanding and generation tasks. They aim to train unified multimodal architectures in an end-to-end manner to unlock more powerful capabilities and emerge potential intelligence. 2) \textbf{Unification of task formats}: These methods usually connect a generation decoder to an existing multimodal understanding foundation model via specific bridging mechanisms, thereby enabling multimodal generation and complex instruction-based editing. Representative works include MetaQuery~\cite{pan2025transfer}, OminiGen2~\cite{wu2025omnigen2}, and \textit{etc.} 3) \textbf{Unification of learning paradigms}: These approaches (such as EMU3~\cite{wang2024emu3} and D-DiT~\cite{li2025dual}) focus on optimization consistency in the unified model, \textit{i.e.}, employing the same learning paradigm {(\textit{e.g.}, next-token prediction or denoising diffusion loss)} for both generation and understanding tasks.

\begin{figure}[p]
    \centering
    \includegraphics[width=0.95\textwidth]{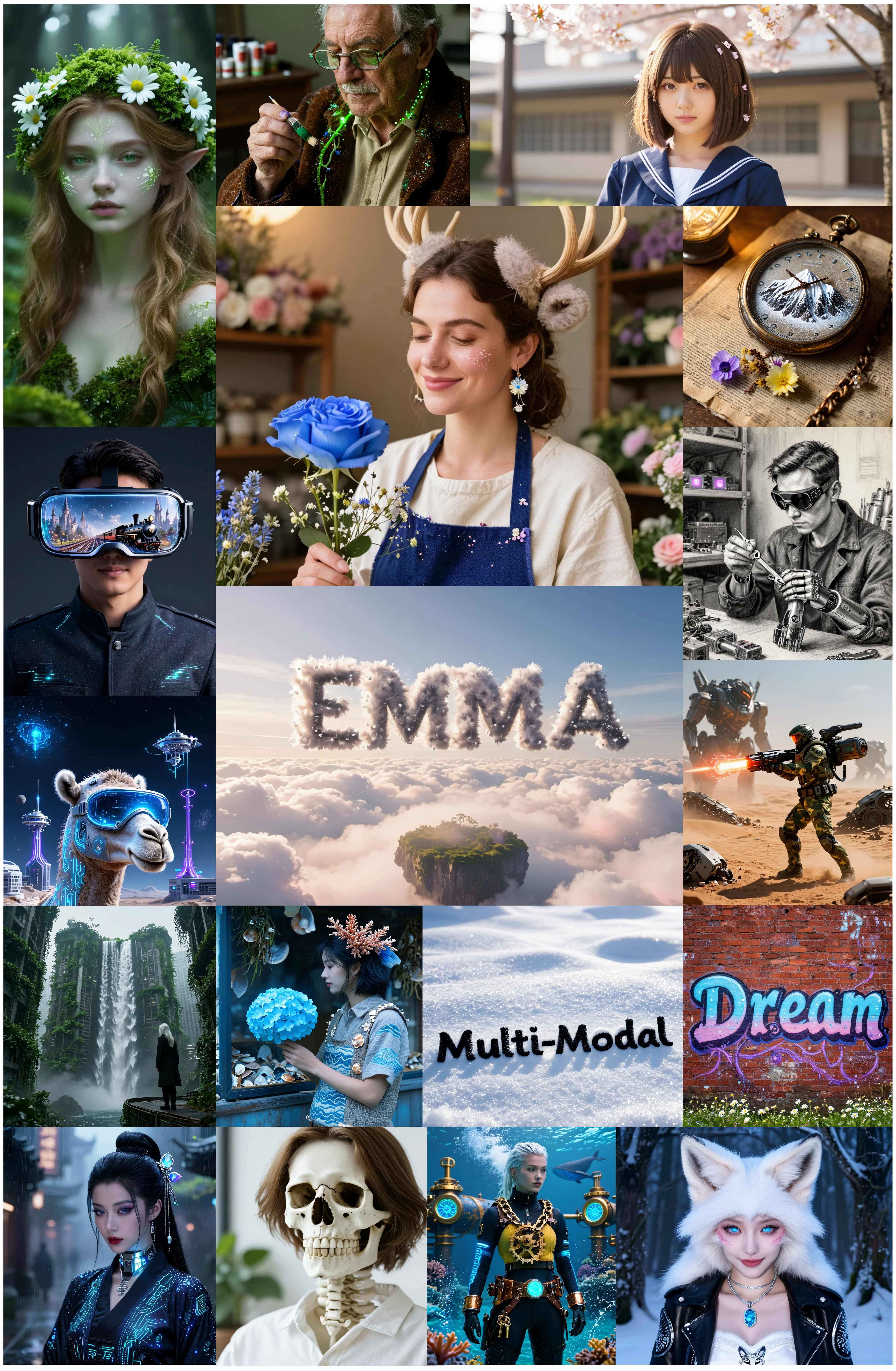}
    \caption{\textbf{Showcase of EMMA's text-to-image generation capability.}} 
    \label{fig:t2i_vis}
\end{figure}

The above technical directions have not yet converged, and each exhibiting distinct advantages and limitations. In this work, EMMA is proposed to solve the inherent issues in the "unification of architecture formats". Specifically, EMMA first uses a higher-compression autoencoder (DCAE~\cite{chen2024deep}, 32x compression), that aligns the compression ratio in the visual understanding branch. This alignment enables the training balances between understanding and generation tasks within the same architecture. Furthermore, benefiting from the same compression ratio used in both the understanding and generation branches, visual tokens extracted from these branches can be concatenated along the channel dimension instead of the token level like BAGEL~\cite{deng2025BAGEL}. This strategy effectively reduces visual context tokens while maintaining the information fusion and complementarity between understanding and generation tasks. In addition, EMMA incorporates a shared-and-decoupled mechanism in its network design. Unlike BAGEL, EMMA introduces cross-task parameter sharing at shallow layers to facilitate mutual improvements between multimodal tasks, while employing parameter decoupling at deeper layers to meet the distinct modeling requirements of understanding and generation. Inspired by Mixpert~\cite{he2025mixpert}, EMMA also integrates a mix-of-experts (MoE) strategy into the visual understanding encoder to improve its perceptual ability across various types of input images.

Using the above designs, EMMA significantly surpasses recent unified multimodal models in extensive multimodal understanding, generation, and editing benchmarks~\cite{lu2024mathvista,masry2022chartqa,ghosh2023geneval,liu2025step1x}, despite using relatively fewer training data and parameters than previous approaches~\cite{deng2025BAGEL,liao2025mogao}. As shwon in Table~\ref{tab:umm_comparison}, EMMA achieves a score of {73.0 on MMVet~\cite{mmvet} and 0.93 on GenEval~\cite{ghosh2023geneval}} based on 4B LLM, compared to the larger unified model BAGEL-7B achieving with 67.2 on MMVet and 0.88 on GenEval. Notably, EMMA only needs $20\%$ visual context tokens compared to BAGEL while handling cross-model interaction generation tasks (\textit{e.g.} image editing). Moreover, compared to the recent open-source state-of-the-art multimodal understanding models, such as InternVL3.5~\cite{wang2025internvl3} and Qwen3-VL~\cite{qwen3vl}, EMMA also achieves competitive results on multiple understanding benchmarks~\cite{mathew2021docvqa,masry2022chartqa,lu2024mathvista}. In addition, EMMA also demonstrates superior performance over Qwen-Image~\cite{wu2025qwen} on GenEval~\cite{ghosh2023geneval} (0.91 \textit{vs.} 0.87 without using prompt rewriting or reinforcement learning strategies). More visualization results are shown in Figure~\ref{fig:t2i_vis}.

\begin{table}[t]
    \centering
     \caption{\textbf{Comparison between different unified multimodal approaches across multimodal understanding, text-to-image generation, and image editing benchmarks.} \textit{Params} represents the number of parameters of utilized LLM. $\dagger$ denotes the methods using prompt rewriting. More results are shown in Section~\ref{sec:experiments}.}
    \resizebox{1.0\linewidth}{!}{
        \begin{tabular}{l|c|ccc|cc|c}
        \toprule
        \textbf{Model}& \textbf{Params}  & \multicolumn{3}{c|}{\textbf{Understanding}} & \multicolumn{2}{c|}{\textbf{T2I Generation}} & {\textbf{Image Editing}} \\

        & & MMBench~\cite{liu2024mmbench} & MMMU~\cite{yue2024mmmu} & MMVet~\cite{mmvet} & GenEval~\cite{ghosh2023geneval} & DPG-Bench~\cite{dpgbench} & GEdit-Bench-EN~\cite{liu2025step1x} \\
        \midrule
        Janus-Pro~\cite{chen2025januspro} & 7B & 79.2 & 41.0 & 50.0 & 0.80 & 84.19  & - \\
        Mogao~\cite{liao2025mogao} & 7B & 75.0 & 44.2 & - & 0.89 & 84.33  & - \\
        OmniGen2~\cite{wu2025omnigen2} & 3B & 79.1 & 53.1 & 61.8 & 0.80 / $0.86^{\dagger}$ & 83.57  & 6.42 \\
        BLIP3-o~\cite{chen2025blip3o} & 7B & 83.5 & 58.6 & 66.6 & $0.84^{\dagger}$ & 81.60 & - \\
        UniWorld-V1~\cite{lin2025uniworld} & 7B & 83.5 & 58.6 & 67.1 & 0.80 / $0.84^{\dagger}$ & 81.38  & 4.85  \\
        BAGEL~\cite{deng2025BAGEL} & 7B & 85.0 & 55.3 & 67.2 & 0.82 / {0.88}$^{\dagger}$ & 85.07 & 6.52 \\
        \midrule
        
        EMMA & 4B & \underline{\textbf{85.8}} & \underline{\textbf{62.5}} & 
        \underline{\textbf{73.0}}&  \underline{\textbf{0.91}} / \underline{\textbf{0.93$^{\dagger}$}} & \underline{\textbf{85.63}}  & \underline{\textbf{6.53}} \\

        \bottomrule
        \end{tabular}
    }
    \label{tab:umm_comparison}
\end{table}

The contributions of this work are summarized as follows:
\begin{itemize}
    \item \textbf{Efficiency.} Through a high-compression autoencoder and channel-wise concatenation mechanism, EMMA significantly reduces the visual tokens (\textit{e.g.}, reducing by 5x in image editing task) compared to recent unified models like BAGEL, boosting end-to-end training and inference efficiency. 
    \item \textbf{Performance.} Benefiting from the shared-and-decoupled architecture design and billions of training data, EMMA achieves superior performances compared to recent unified multimodal models.
    \item \textbf{Potential.} EMMA shows competitive results, achieved by a unified architecture training with the end-to-end manner, compared to understanding and generation state-of-the-arts. The above results further highlight its potential for cross-modal tasks.
\end{itemize}

\section{Method}
The overall architecture of EMMA is shown in Figure~\ref{fig:architecture}. The core of EMMA lies in its visual encoder module and network architecture. The encoder ensures effective visual information fusion, while the network architecture enables cross-task parameter sharing and meets the distinct modeling requirements of understanding (semantics modeling) and generation (semantics and high-frequency details modeling~\cite{ma2025deco}) tasks. The details of these two modules are shown below.

\subsection{Visual Encoder}

\textbf{Choice of Visual Understanding and Generation Encoder.} For unified multimodal architectures, there are usually two types of visual encoder: understanding encoder (denoted as \textit{Und-Enc}) and generation encoder (denoted as \textit{Gen-Enc}). For \textit{Und-Enc}, recent works generally adopt SigLIP~\cite{siglip,siglip2} to encode images as visual tokens and extract the corresponding semantic information, followed by a 2x2 pixel shuffle strategy~\cite{internvl15} to further reduce these tokens to one quarter before feeding them into the Large Language Model (LLM). In this work, we directly utilize \textit{SigLIP2-so400m-patch16-512}~\cite{siglip2} as \textit{Und-Enc}. Meanwhile, we extend its capability to support native resolutions of input images by interpolating the positional embeddings. As a result, \textit{Und-Enc}, through the patchfy operation in SigLIP2 and pixel shuffling strategy, achieves a 32× compression ratio of the input image. For example, a 1024×1024 resolution image is compressed into 1024 visual tokens. For \textit{Gen-Enc}, we employ a high-compression autoencoder, DCAE~\cite{dcae}, with a 32× compression ratio. Compared to other unified architectures that generally adopt the autoencoder (AE) with an 8x compression ratio (\textit{e.g.}, the AE in FLUX~\cite{flux2024}) and 2x2 token merging strategy, EMMA only requires 1/4 visual tokens for the generation task. Experimental results in Section~\ref{sec:experiments} demonstrate that using the above high-compression AE still yields competitive generation quality.

\noindent
\textbf{Fusion of Visual Information.} Furthermore, since both \textit{Und-Enc} and \textit{Gen-Enc} utilize the same compression ratio (32×), EMMA can directly perform channel-wise concatenation of the corresponding visual tokens, rather than the token-wise concatenation used in previous approaches like BAGEL~\cite{deng2025BAGEL}. This allows EMMA to effectively fuse semantic information (from \textit{Und-Enc}) with detailed information (from \textit{Gen-Enc}) without increasing the total number of visual tokens, thereby supporting more efficient training and inference of unified models.
Consequently, compared to recent state-of-the-art unified models (\textit{e.g.}, BAGEL~\cite{deng2025BAGEL}), EMMA significantly reduces the visual tokens (e.g., reducing by 5x in image editing task), while still achieving superior performances across various multimodal understanding, generation, and editing benchmarks~\cite{yue2024mmmu,ghosh2023geneval,liu2025step1x}.

\begin{figure}[t]
    \centering
    \includegraphics[width=\textwidth]{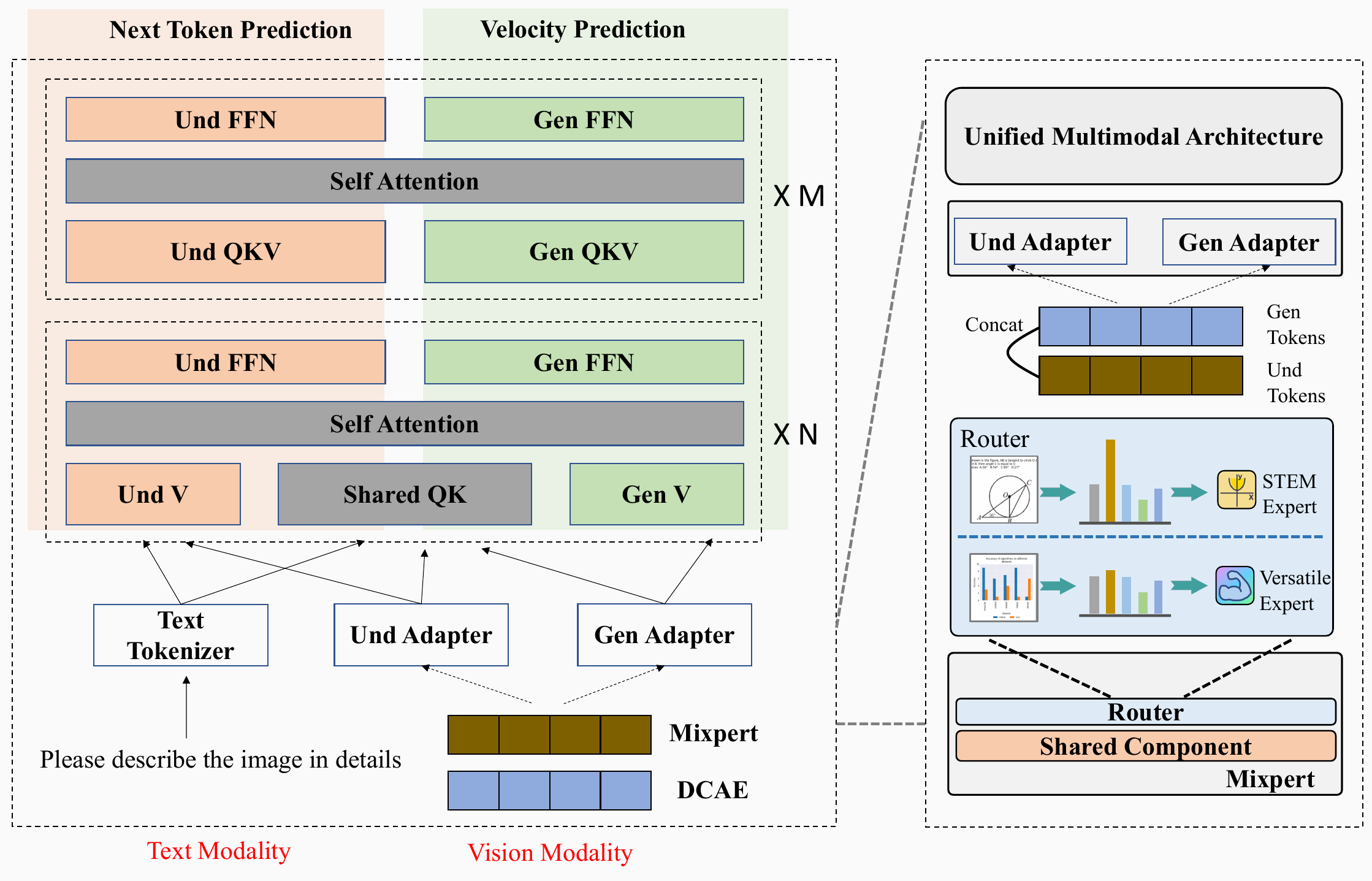}
    \caption{\textbf{The overall architecture of our EMMA.} "Und Adapter" and "Gen Adapter" represent understanding adapter and generation adapter for projecting visual tokens into unified multimodal architecture, respectively.    } 
    \label{fig:architecture}
\end{figure}

\noindent
\textbf{Mixture of Experts.} In addition, the types of input images in multimodal understanding tasks are highly different. Inspired by Mixpert~\cite{mixpert}, EMMA further extends SigLIP2 into a mixture-of-experts architecture. As shown in Figure~\ref{fig:architecture}, EMMA additionally introduces a \textit{STEM} (science, technology, engineering, and math) expert to process \textit{STEM} images, with a router module dynamically selecting the appropriate expert for each input. Specifically, when the input image is identified as \textit{STEM} data by the router, it would be fed into the \textit{STEM} expert; otherwise, it defaults to the versatile expert. Given that \textit{STEM} data appear almost exclusively in multimodal understanding tasks, the weights of \textit{STEM} expert are first initialized from the versatile expert and only trained in the last tuning phase with the mutlimodal understanding data ({the training details are presented in Section~\ref{sec:training_details}}). {Experimental results in Section~\ref{sec:experiments}} demonstrate that by adding only approximately 50M additional parameters and few fine-tuning steps, EMMA improves performance on multimodal understanding benchmarks while preserving original generation and editing capabilities.

\subsection{Network Architecture Design}
\textbf{Network Design.} As shown in Figure~\ref{fig:architecture}, EMMA consists of a shared-and-decoupled network. This design is mainly motivated by understanding (denoted as \textit{Und}) tasks focusing on semantics modeling and generation (denoted as \textit{Gen}) tasks focusing on semantics and high-frequency information (details) modeling, thereby there are both common and independent characteristics. Considering that deeper layers exhibit stronger relevance to specific tasks, EMMA adopts a mechanism of sharing shallow layers while keeping deeper layers independent. Additionally, in shallow layes, EMMA still maintains a portion of parameters (\textit{e.g.}, the value projection layers) as task-specific module to ensure task independence. Moreover, cross-modal interactions are achieved through self-attention layers. In particular, the weights of all layers in both \textit{Und} and \textit{Gen} branches are initialized from Qwen3-4B~\cite{yang2025qwen3}. 

\noindent
\textbf{Data Processing and Optimization Objective.} Therefore, for each task, textual inputs are processed through the \textit{Und} branch. Visual inputs first undergo semantic feature and details extraction via the modified Mixpert and DCAE. These extracted information are then fused through channel concatenation, and then feed into \textit{Und} or \textit{Gen} branch according to the corresponding tasks. As for \textit{Und} tasks, EMMA utilizes the next-token prediction mechanism to guide overall learning. For \textit{Gen} tasks, EMMA utilizes flow matching with velocity prediction. Notably, before feeding visual tokens into the LLM, the 2D positional encoding is applied to incorporate spatial priors. Subsequently, all textual and visual tokens are treated uniformly and processed using the positional embedding of 1D RoPE~\cite{rope}.

\noindent
\textbf{Attention Strategy.} Following Transfusion~\cite{zhou2024transfusion}, EMMA adopts a hybrid attention strategy. Specifically, pure casual masking is utilized to both textual and visual tokens in \textit{Und} tasks. As for \textit{Gen} tasks, textual tokens are still restricted to attend previous tokens, and visual tokens can attend both of previous tokens and other visual tokens within the same image.


\section{Data Construction}
\label{sec:data_construction}
EMMA is trained with end-to-end manner, and the training data consists of mutlimodal understanding data, text-to-image generation data and instruction editing data. {The detailed statistics are presented in Table~\ref{tab:training_data_details}}.

\noindent
\textbf{Multimodal Understanding Data (I2T).} This dataset is primarily constructed from six types of data: alignment data, pre-training (PT) data, supervised fine-tuning (SFT) data, quality tuning (QT) data, \textit{STEM} expert tuning data (ET), and router tuning data (RT). Alignment data are used to align the extracted visual tokens with LLM, and this work directly uses LLaVA-558K~\cite{liu2024improved}. PT data mainly consists of text-image pairs, sourced predominantly from large-scale open-source datasets such as LAION~\cite{schuhmann2021laion}, as well as internal data. Additionally, re-captioning models~\cite{internvl15,bai2025qwen2} are employed to regenerate captions for these images. SFT data are largely composed of image-question-answer triplets. These are mainly derived from large-scale, high-quality open-source datasets (including LLaVA-OneVision-Data~\cite{li2024llava}, FineVision~\cite{wiedmann2025finevision} and \textit{etc}.), along with the internal constructed datasets. SFT data cover a wide range of types, such as document parsing, chart recognition, optical character recognition, mathematical problem solving, \textit{etc}. QT data are built by selecting higher-quality samples from SFT data and performing balanced sampling across each task. {ET data consist of 15M \textit{STEM} data sourced from SFT data, and RT data are constructed by combining 3M data with both \textit{STEM} and general data.}

\noindent
\textbf{Text-to-Image Generation Data (T2I).} This dataset is composed of three data types: PT, SFT, and QT data. PT data are mainly filtered from large-scale open-source datasets (such as LAION~\cite{schuhmann2021laion}) as well as internal datasets, based on aesthetic quality. SFT data are selected according to the filter criterias such as image resolution (1K resolution and above) and aesthetic score, with the balanced sampling between general images and portrait images. To address the scarcity of text rendering data, we have also synthesized text rendering images with current state-of-the-art generation models~\cite{wu2025qwen}.

\begin{table}
\small
\centering

\caption{\textbf{The training data and hyperparameters of EMMA}. I2T, T2I and IT2I represents the multimodal understanding, text-to-image generation and image editing data, respectively. LR denotes the learning rate.}
    \label{tab:training_data_details}
\resizebox{1.0\linewidth}{!}{
\aboverulesep=5pt
\begin{tabular}{l|cccccc}
\toprule
 & \textbf{Alignment} & \textbf{PT} & \textbf{SFT} & \textbf{QT} & \textbf{ET} & \textbf{RT}\\
\hline

\textbf{Training Data(M)} \\
I2T Data          & $0.56$ & $400$ & $120$ & $1$ & $15$ & $3$ \\
T2I Data          & - & $600$ & $105$ & $0.15$ & - & -  \\
IT2I Data          & - & - & $12$ & $0.35$ & - & -  \\

\midrule
\textbf{Hyperparameters} \\
\textit{Und} resolution      & $(512, 512)$ & $(512, 512)$ & native & native & native & native \\
\textit{Gen} resolution     & - & $(512, 512)$ & $1$K & $1$K & - & - \\
LR ($\times10^{-4}$)  & $10.0$ & $1.0$ & $0.4$ & $0.1$ & $0.04$ & $1.0$ \\
LR scheduler   & Cosine & Constant & Constant & Cosine& Cosine & Constant \\
Optimizer       & \multicolumn{6}{c}{AdamW ($\beta_1=0.9$, $\beta_2=0.999$, $\epsilon=1.0 \times 10^{-8}$)} \\

\bottomrule
\end{tabular}
}
\end{table}

\begin{figure}[t]
    \centering
    \includegraphics[width=\textwidth]{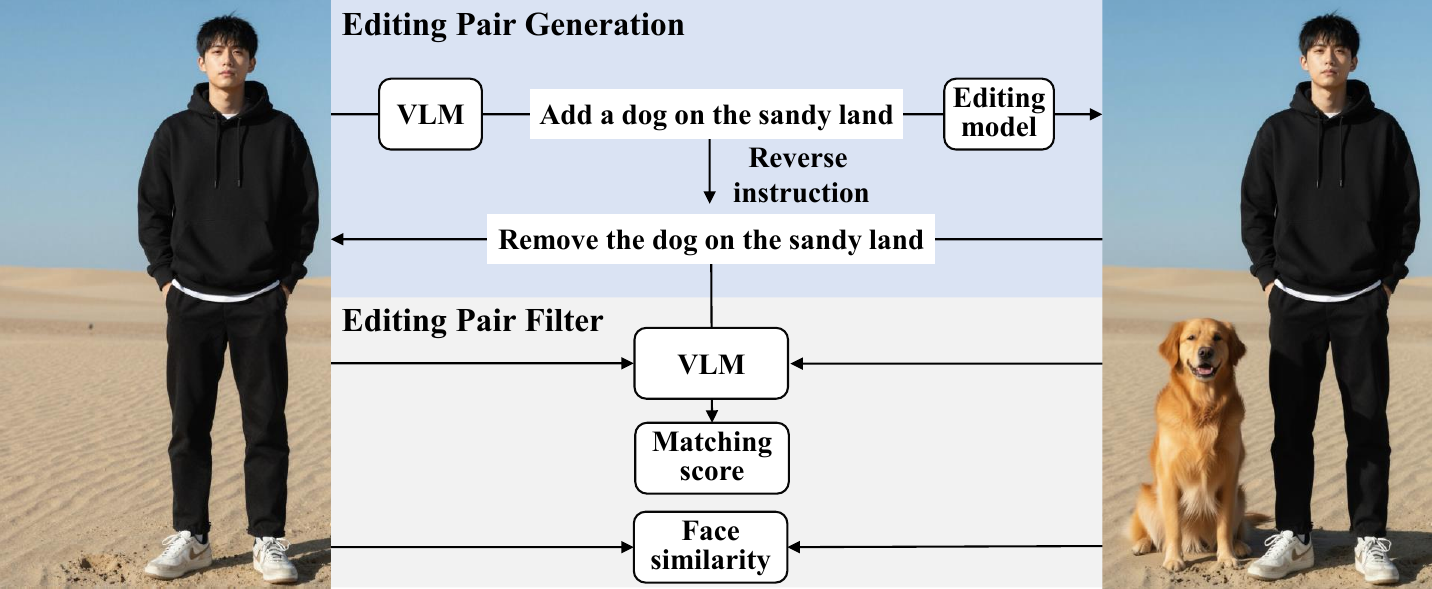}
    \caption{\textbf{Overview of the general editing data construction pipeline.} The data pipeline consists of two parts. The first is to generate image editing pairs: firstly, we use a VLM to generate editing instruction for the input image, then use an image editing model to generate the edited image as well as reversing the editing instruction to obtain the reversed pair. The second part is to filter the image editing pairs: we use a VLM to determine whether the edited image follows the editing instruction, and if the image contains portraits, we further filter the pair using face similarity.} 
    \label{fig:data_general}
\end{figure}

\begin{figure}[t]
    \centering
    \includegraphics[width=\textwidth]{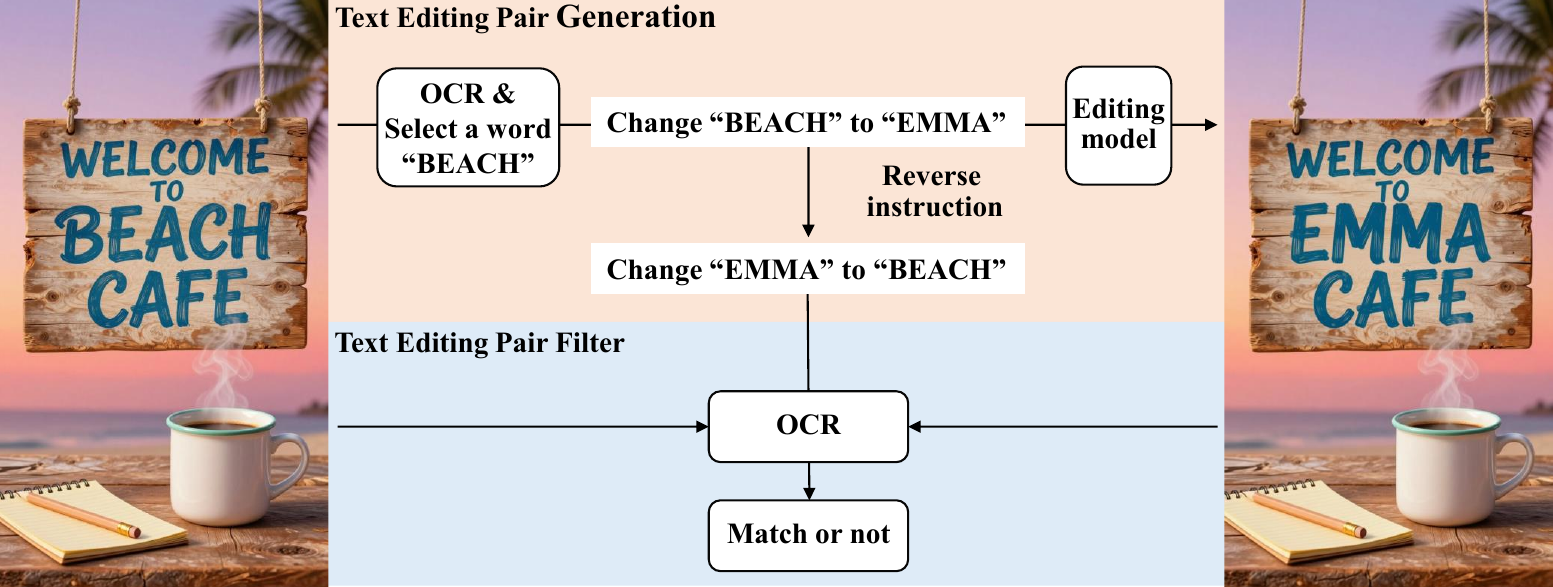}
    \caption{\textbf{Overview of the text editing data construction pipeline.} The data pipeline consists two parts. The first is to generate text-edited image pairs: firstly, text detection is performed on the input image to extract text information, and then one or more words are randomly selected for replacement or removal, while generating the corresponding editing instruction. We further use a image editing model to produce the edited image. The second part is to filter the image pairs, where the OCR model is used to determine whether the edited image follows the instruction or not.} 
    \label{fig:data_text}
\end{figure}

\noindent
\textbf{Image Editing Data (IT2I).} Currently, several high-quality image editing datasets have been released, such as X2I2~\cite{wu2025omnigen2} and {OmniEdit~\cite{wei2025omniedit}}. In addition to leveraging these existing open-source datasets, we also use state-of-the-art editing models (\textit{e.g.}, Qwen-Image-Edit~\cite{wu2025qwen}) to synthesize a number of high-quality editing pairs. The construction and filtering pipeline is illustrated in Figure~\ref{fig:data_general}. The above pipeline is designed to produce data covering various editing types, including object addition, removal, replacement, background transfer, tone transfer, and virtual try-on. Especially, we also construct a data pipeline for text editing tasks, and the details are shown in Figure~\ref{fig:data_text}. The final SFT data are constructed by combining the open-source datasets with our synthesized editing pairs. Furthermore, the synthesized editing data (the reference images are sourced from T2I QT data) are regarded as the final editing QT data. \textbf{It should be noted that although the recently released GPT-Image-Edit-1.5M~\cite{wang2025gpt} can significantly improve performance on image editing benchmarks, it severely disrupts the character consistency.} Therefore, we believe that this dataset is harmful to region-based editing tasks and excludes it from our training corpus. This observation also highlights the need for more accurate evaluation metrics in image editing (such as subject consistency), which is left for future work.

\section{Training Details}
\label{sec:training_details}

\noindent
\textbf{Stage 0: Alignment.} This stage aims to align the visual tokens with unified model. Therefore, the parameters of visual encoders and unified model are frozen, and only the adapter in \textit{Und} branch is tuned. The input image is fixed at 512x512. {The learning rate is set as $1e^{-3}$}. As for \textit{Gen} branch, the adapter is randomly initialized and left for training during the pre-training stage. 

\noindent
\textbf{Stage 1: Pre-Training (PT).} In this stage, all parameters of EMMA (expect for visual generation encoder, \textit{i.e.}, DCAE) are trained. The input image is set as 512x512 for both \textit{Und} and \textit{Gen} branches, and their corresponding batch sample ratio is set as 1:1. {The learning rate is set as $1e^{-4}$}.

\noindent
\textbf{Stage 2: Supervised Finetuning (SFT).} Consistent with PT stage, all parameters are still tuned except for DCAE in this stage. Differently, \textit{Und} branch supports training with the native resolution of input images, while \textit{Gen} branch supports training with input images of which are scaled to the nearest pre-defined bucket size of 1K resolution based on their original aspect ratios. 
Furthermore, after training with one epoch on the initial I2T and T2I SFT data, EMMA is further fine-tuned by using a balanced sampling strategy. Specifically, for T2I SFT data, {we select $\sim$50M samples with a 1:1 ratio between portrait and general images, while ensuring the balances across aspect ratios and incorporating aesthetic-based ranking strategy. Similarly, in I2T SFT data, we sample $\sim$50M instances with a 1:1 ratio between \textit{STEM} and general categories.} Towards the end of its training, we further incorporate IT2I SFT data and train EMMA with a balanced mixture ratio of 1:1:1 between all three tasks. 

\noindent
\textbf{Stage 3: Quality Tuning (QT).} Consistent with SFT stage, we conduct joint training among T2I, I2T, and IT2I tasks with a balanced batch ratio of 1:1:1. {The initial learning rate is set as $1e^{-5}$ in this stage.}

\noindent
\textbf{Stage 4: STEM Expert and Router Tuning (ET\&RT).} In these two stages, only the parameters of \textit{STEM} expert or router module are trained and other parameters are frozen. {The STEM expert is trained with 12M \textit{STEM} data, and the router is trained with the specially chosen 3M data as described in Section~\ref{sec:data_construction}.} {The initial learning rate is set as $4e^{-6}$ in ET stage and $1e^{-4}$ in RT stage.}

\section{Experiments}
\label{sec:experiments}
\subsection{Comparisons on Understanding Tasks}
We compare EMMA against current state-of-the-art methods on multiple multimodal understanding benchmarks. As shown in Table \ref{tab:und}, EMMA significantly outperforms existing unified multimodal architectures. For example, EMMA, built on 4B LLM, can significantly surpass BAGEL-7B by a notable margin, {\textit{e.g.}, achieving with a 5.8\% improvement on the MMVet benchmark.} 

Furthermore, reconstructing the vision encoder with a mixture-of-experts (MoE) design further improves the accuracy of EMMA in understanding benchmarks, yielding an average gain of 0.4\% across 11 evaluation datasets. In particular, EMMA remains highly competitive results compared to current multimodal understanding state-of-the-arts, \textit{e.g.}, it outperforms Qwen3-VL by an average of 0.4\% and exceeds InternVL3.5 by an average of 2.6\% on 11 evaluation datasets. The above results clearly demonstrate the effectiveness of EMMA.

\begin{table}
  \centering
  \caption{\textbf{Comparisons with state-of-the-arts on multimodal understanding benchmarks.} \textit{Params} represents the number of parameters of utilized LLM. $\S$ denotes the results tested by ourself with {VLMEvalKit}~\cite{duan2024vlmevalkit} using the officially released instruct-version checkpoint. CQA, DVQA, TVQA, IVQA, OCRB, MMB. MVista, MMS and MMV represents ChartQA~\cite{masry2022chartqa}, DocVQA~\cite{mathew2021docvqa}, TextVQA~\cite{textvqa}, InfoVQA~\cite{infovqa}, OCRBench~\cite{liu2024ocrbench}, MMBench~\cite{liu2024mmbench}, MathVista~\cite{lu2023mathvista}, MMstar~\cite{mmstar} and MMVet~\cite{mmvet}, respectivelly. \textit{W/O} represents the method without extending SigLIP2\cite{siglip2} into a mixture-of-experts architecture.}
  \label{tab:und}
  \centering
  \small
  \resizebox{1.0\linewidth}{!}{
  \begin{tabular}{l|c|ccccccccccc|l}
    \toprule
    \textbf{Method} & \textbf{Params} & \textbf{CQA} & \textbf{DVQA} &\textbf{TVQA} &\textbf{IVQA}  & \textbf{OCRB} & \textbf{MMB} &\textbf{MMMU} &\textbf{MVista} &\textbf{AI2D} & \textbf{MMS} & \textbf{MMV}  & \textbf{Avg} \\
    \midrule
    LLaVA-OV-1.5~\cite{an2025llavaov15}& 4B & 87.1 & 94.4 & - & 76.1 & 80.0 & {84.2} & 52.7 & 67.9 & 83.6 & 64.9 & - & -\\
    InternVl-3.5$^\S$~\cite{wang2025internvl3}& 4B & 86.3 & 91.9 & 77.8 & 77.5 & 81.7 & 81.4 & 60.0 & 68.4 & 82.0 & 64.0 & \underline{\textbf{76.1}} & 77.0\\
    Qwen3-VL$^\S$~\cite{qwen3vl}& 4B & 82.8 & 95.2 & \underline{\textbf{80.9}} & \underline{\textbf{79.9}} & 88.1 & 83.6 & \underline{\textbf{67.4}} & 73.7 & \underline{\textbf{83.9}} & \underline{\textbf{69.8}} & 66.2 & 79.2\\
    \midrule
    Janus Pro~\cite{chen2025januspro} & 7B & - & - & - & - & - & 79.2 & 41.0 & - & - & - & 50.0 & -\\
    EMU3~\cite{wang2024emu3}& 8B & 68.6 & 76.3 & 64.7 & 43.8 & 68.7 & 58.5 & 31.6 & - & 70.0 & - & 37.2 & -\\
    VILA-U~\cite{wu2024vilau}& 7B & - & - & 60.8 & - & - & - & - & - & - & - & 33.5 & -\\
    ILLUME~\cite{wang2025illume}& 7B & 66.7 & 76.0 & - & 45.5 & 66.9 & 75.1 & 38.2 & - & 71.4 & - & 37.0 & -\\
    MUSE-VL~\cite{xie2025musevl}& 7B & - & - & - & - & - & 72.1 & 39.7 & 51.3 & 69.8 & 49.6 & - & -\\
    OminiGen2~\cite{wu2025omnigen2}& 3B & - & - & - & - & - & 79.1 & 53.1 & - & - & - & 61.8 & -\\
    UniWolrd-V1~\cite{lin2025uniworld}& 7B & - & - & - & - & - & 83.5 & 58.6 & - & - & - & 67.1 & -\\
    BLIP3-o~\cite{chen2025blip3o}& 7B & - & - & - & - & - & 83.5 & 58.6 & - & - & - & 66.6 & -\\
    Mogao~\cite{liao2025mogao}& 7B & - & - & - & - & - & 75.0 & 44.2 & - & - & - & - & -\\
    Show-o2~\cite{xie2025showo2}& 7B & - & - & - & - & - & 79.3 & 48.9 & - & 78.6 & 56.6 & - & -\\
    BAGEL~\cite{deng2025BAGEL}& 7B & - & - & - & - & - & 85.0 & 55.3 & 73.1 & - & - & 67.2 & -\\
    \midrule
    EMMA \textit{(W/O)}& 4B & 87.8 & 95.5 & 79.0 & 77.4 & 88.4 & 85.6 & 62.0 & 75.4 & 83.6 & 64.2 & 72.5 & 79.2\\
    EMMA& 4B & \underline{\textbf{88.0}} & \underline{\textbf{95.9}} & 79.5 & 77.7 & \underline{\textbf{89.0}} & \underline{\textbf{85.8}} & 62.5 & \underline{\textbf{75.8}} & \underline{\textbf{83.9}} & 64.8 & 73.0 & \underline{\textbf{79.6}}\\
    \bottomrule
  \end{tabular}
  }
\end{table}

\subsection{Comparisons on Text-to-Image Generation Tasks}
To further evaluate the text-to-image generation capability of EMMA, we conduct comprehensive comparisons on GenEval~\cite{liu2025step1x} and DPG-Bench~\cite{dpgbench} against both unified multimodal architectures and current text-to-image (T2I) generation approaches. As shown in Table~\ref{tab:geneval} and Table~\ref{tab:dpg}, EMMA significantly outperforms existing unified approaches in T2I generation benchmarks. For example, on GenEval, EMMA achieves a score of 0.91 without prompt rewriting and reinforcement learning, surpassing BAGEL-7B (0.82) and Qwen-Image (0.87), despite the latter having a larger model size (4B \textit{vs.} 20B). It is worth noting that, to our best knowledge, EMMA is the first to reach a score of 0.91 on GenEval without relying on prompt rewriting or reinforcement learning. This further demonstrates the benefits of unified architecture for T2I generation task. {As shown in Figure~\ref{fig:t2i_vis}, EMMA is capable of generating high-quality images following the given textual instruction.}

\begin{table}
    \centering
    \caption{\textbf{Quantitative evaluations of the text-to-image generation capacity on GenEval~\cite{ghosh2023geneval}}. $\dagger$ denotes the methods using prompt rewriting.}
    \label{tab:geneval}
    \resizebox{1.0\linewidth}{!}{
        \begin{tabular}{l|cccccc|c}
            \toprule
            \textbf{Method} & \textbf{Single object} & \textbf{Two object} & \textbf{Counting} & \textbf{Colors} & \textbf{Position} & \textbf{Color attribution} & \textbf{Overall}$\uparrow$ \\
            \midrule
            LUMINA-Next~\cite{zhuo2024lumina} & 0.92 & 0.46 & 0.48 & 0.70 & 0.09 & 0.13 & 0.46 \\
            SDXL~\cite{podell2023sdxl} & 0.98 & 0.74 & 0.39 & 0.85 & 0.15 & 0.23 & 0.55 \\
            FLUX.1-dev~\cite{flux2024} & 0.99 & 0.81 & 0.79 & 0.74 & 0.20 & 0.47 & 0.67 \\
            FLUX.1-dev$^{\dagger}$~\cite{flux2024} & 0.98 & 0.93 & 0.75 & 0.93 & 0.68 & 0.65 & 0.82 \\
            SD3-medium~\cite{sd3} & 0.99 & 0.94 & 0.72 & 0.89 & 0.33 & 0.60 & 0.74 \\
            SANA-1.5~\cite{xie2025sana} & 0.99 & 0.93 & 0.86 & 0.84 & 0.59 & 0.65 & 0.81 \\
            HiDream-I1-Full~\cite{cai2025hidream} & \underline{\textbf{1.00}} & 0.98 & 0.79 & 0.91 & 0.60 & 0.72 & 0.83 \\
            Qwen-Image~\cite{wu2025qwen} & 0.99 & 0.92 & 0.89 & 0.88 & 0.76 & 0.77 & 0.87 \\
            Qwen-Image-RL~\cite{wu2025qwen} & \underline{\textbf{1.00}} & 0.95 & \underline{\textbf{0.93}} & 0.92 & \underline{\textbf{0.87}} & 0.83 & 0.91 \\
            \midrule
            ILLUME~\cite{wang2025illume} & 0.99 & 0.86 & 0.45 & 0.71 & 0.39 & 0.28 & 0.61 \\
            $\text{Emu3-Gen}^{\dagger}$~\cite{wang2024emu3} & 0.99 & 0.81 & 0.42 & 0.80 & 0.49 & 0.45 & 0.66 \\
            Show-o~\cite{xie2024showo} & 0.98 & 0.80 & 0.66 & 0.84 & 0.31 & 0.50 & 0.68 \\
            Janus Pro~\cite{chen2025januspro} & 0.99 & 0.89 & 0.59 & 0.90 & 0.79 & 0.66 & 0.80 \\
            $\text{MetaQuery-XL}^{\dagger}$~\cite{pan2025transfer} & - & - & - & - & - & - & 0.80 \\
            UniWorld-V1~\cite{lin2025uniworld} & 0.99 & 0.93 & 0.79 & 0.89 & 0.49 & 0.70 & 0.80 \\
            $\text{BLIP3-o}^{\dagger}$~\cite{chen2025blip3o} & - & - & - & - & - & - & 0.84 \\
            {OminiGen2}~\cite{wu2025omnigen2} & \underline{\textbf{1.00}} & 0.95 & 0.64 & 0.88 & 0.55 & 0.76 & 0.80 \\            
            {OminiGen2}$^{\dagger}$~\cite{wu2025omnigen2} & 0.99 & 0.96 & 0.74 & \underline{\textbf{0.98}} & 0.71 & 0.75 & 0.86 \\
            BAGEL~\cite{deng2025BAGEL} & 0.99 & 0.94 & 0.81 & 0.88 & 0.64 & 0.63 & 0.82 \\
            $\text{BAGEL}^{\dagger}$~\cite{deng2025BAGEL} & 0.98 & 0.95 & 0.84 & 0.95 & 0.78 & 0.77 & 0.88 \\
            Mogao~\cite{liao2025mogao} & \underline{\textbf{1.00}} & 0.97 & 0.83 & 0.93 & 0.84 & 0.80 & 0.89 \\
            \midrule
            \textbf{EMMA} & \underline{\textbf{1.00}} & 0.98 & 0.83 & 0.96 & 0.83 & 0.85 & 0.91 \\            \textbf{EMMA}$^{\dagger}$ & \underline{\textbf{1.00}} & \underline{\textbf{0.99}} & 0.87 & \underline{\textbf{0.98}} & 0.86 & \underline{\textbf{0.87}} & \underline{\textbf{0.93}} \\
            \bottomrule
        \end{tabular}
    }
    \label{tab:geneval}
\end{table}

\begin{table}[t]
    \centering
    \caption{\textbf{Quantitative evaluations of the text-to-image generation capacity on DPG-Bench~\cite{dpgbench}.}}
    \label{tab:dpg}
\resizebox{0.95\linewidth}{!}{
    \begin{tabular}{l|ccccc|c}
        \toprule
        \textbf{Method} & \textbf{Global} & \textbf{Entity} & \textbf{Attribute} & \textbf{Relation} & \textbf{Other} & \textbf{Overall}$\uparrow$ \\
        \midrule
        SD v1.5~\cite{sd15} & 74.63 & 74.23 & 75.39 & 73.49 & 67.81 & 63.18 \\
        LUMINA-Next~\cite{zhuo2024lumina} & 82.82 & 88.65 & 86.44 & 80.53 & 81.82 & 74.63 \\
        SDXL~\cite{podell2023sdxl} & 83.27 & 82.43 & 80.91 & 86.76 & 80.41 & 74.65 \\ 
        SD3-medium~\cite{sd3} & 87.90 & 91.01 & 88.83 & 80.70 & 88.68 & 84.08 \\
        FLUX.1-dev~\cite{flux2024} & 74.35 & 90.00 & 88.96 & 90.87 & 88.33 & 83.84 \\
        SANA-1.5~\cite{xie2025sana} & - & - & - & - & - & 84.7 \\
        GPT Image 1 [High]~\cite{gptimage} & 88.89 & 88.94 & 89.84 & 92.63 & 90.96 & 85.15 \\
        HiDream-I1-Full~\cite{cai2025hidream} & 76.44 & 90.22 & 89.48 & 93.74 & 91.83 & 85.89 \\
        Qwen-Image~\cite{wu2025qwen} & \underline{\textbf{91.32}} & 91.56 & \underline{\textbf{92.02}} & \underline{\textbf{94.31}} & \underline{\textbf{92.73}} & \underline{\textbf{88.32}} \\
        
        \midrule
        Show-o~\cite{xie2024showo} & 79.33 & 75.44 & 78.02 & 84.45 & 60.80 & 67.27 \\
        Emu3-Gen~\cite{wang2024emu3} & 85.21 & 86.68 & 86.84 & 90.22 & 83.15 & 80.60 \\ 
        UniWorld-V1~\cite{lin2025uniworld} & 83.64 & 88.39 & 88.44 & 89.27 & 87.22 & 81.38 \\
        BLIP3-o~\cite{chen2025blip3o} & - & - & - & - & - & 81.60 \\
        {OmniGen2}~\cite{wu2025omnigen2} & 88.81 & 88.83 & 90.18 & 89.37 & 90.27 & 83.57 \\
        Janus Pro~\cite{chen2025januspro} & 86.90 & 88.90 & 89.40 & 89.32 & 89.48 & 84.19 \\
        Mogao~\cite{liao2025mogao} & 82.37 & 90.03 & 88.26 & 93.18 & 85.40 & 84.33 \\
        BAGEL~\cite{deng2025BAGEL} & 88.94 & 90.37 & 91.29 & 90.82 & 88.67 & 85.07 \\
        \midrule
        EMMA & 91.24 & \underline{\textbf{91.71}} & 90.59 & 92.23 & 90.02 & 85.63 \\
        \bottomrule
    \end{tabular}
    
}
\end{table}

\subsection{Comparisons on Editing Tasks}
We also compare EMMA with current unified architectures on image editing task. As shown in Table~\ref{tab:geneval}, EMMA has a slight advantage over the existing unified multimodal models on GEdit~\cite{liu2025step1x}. This marginal improvement may be attributed to two reasons: 1) {the image-text interaction data used in our approach is relatively limited (12M in EMMA \textit{vs.} 65M in BAGEL);} 2) image editing is a more complex task, and addressing the corner cases should require larger model. \textbf{It is worth noting that in image editing, EMMA only needs 1/5 of the visual tokens for reference images but achieves better results than BAGEL-7B.}

In addition, we have observed the limitation of evaluation metrics in GEdit. Specifically, GEdit relies on a vision-language model to assess editing results, which causes the failure to evaluate subject consistency. Although many recent methods~\cite{wang2025gpt,wang2025lightbagel} incorporate the GPT-Image-Edit-1.5M dataset~\cite{wang2025gpt} (generated via ChatGPT~\cite{gptimage}) to significantly boost GEdit scores, we observed that such data severely disrupt subject consistency. As a result, we exclude this dataset from our final training corpus. This observation also highlights the need for more accurate evaluation criteria for image editing, which is left for future work.

\label{subsec:editing}
\begin{table}[t]
\centering
\caption{\textbf{Comparisons between recent approaches on GEdit-Bench-EN~\cite{liu2025step1x}.} The methods trained with the GPT-Image-Edit-1.5M~\cite{wang2025gpt} are not included.} 
\label{gedit}
\aboverulesep=6pt
\resizebox{0.97\linewidth}{!}{
    \begin{tabular}{l|cccccc}
        \toprule
        \multirow{2}{*}{\textbf{Method}} & \multicolumn{3}{c}{\textbf{GEdit-Bench-EN}} \\
        & \multicolumn{1}{c}{\space \textbf{Semantic Consistency}$\uparrow$} & \multicolumn{1}{c}{\space \textbf{Perceptual Quality}$\uparrow$} & \multicolumn{1}{c}{\space \textbf{Overall Score} $\uparrow$} \\
        \midrule
        Instruct-Pix2Pix~\cite{brooks2023instructpix2pix} & 3.58 & 5.49 & 3.68 \\
        AnyEdit~\cite{yu2025anyedit} & 3.18 & 5.82 & 3.21 \\
        MagicBrush~\cite{zhang2023magicbrush} & 4.68 & 5.66 & 4.52 \\
        OmniGen~\cite{xiao2025omnigen} & 5.96 & 5.89 & 5.06 \\
        FLUX.1 Kontext [Pro]~\cite{flux2024} & 7.02 & 7.60 &  6.56 \\
        Step1X-Edit~\cite{liu2025step1x} & 7.09 & 6.76 & {6.70} \\
        Qwen-Image~\cite{wu2025qwen}  & \underline{\textbf{8.00}} & \underline{\textbf{7.86}} &  \underline{\textbf{7.56}}  \\
        \midrule
        UniWorld-V1~\cite{lin2025uniworld} & 4.93 & {7.43} & 4.85 \\
        {OmniGen2}~\cite{wu2025omnigen2} & {7.16} & 6.77 & 6.41 \\
        BAGEL~\cite{deng2025BAGEL} & {7.36} & 6.83 & 6.52 \\
        \midrule
        EMMA & 7.12 & 6.85 & 6.53 \\
        \bottomrule
    \end{tabular}
    }
\end{table}

\subsection{Emergence of Capabilities}

\begin{figure}[p]
    \centering
    \includegraphics[width=1.0\textwidth]{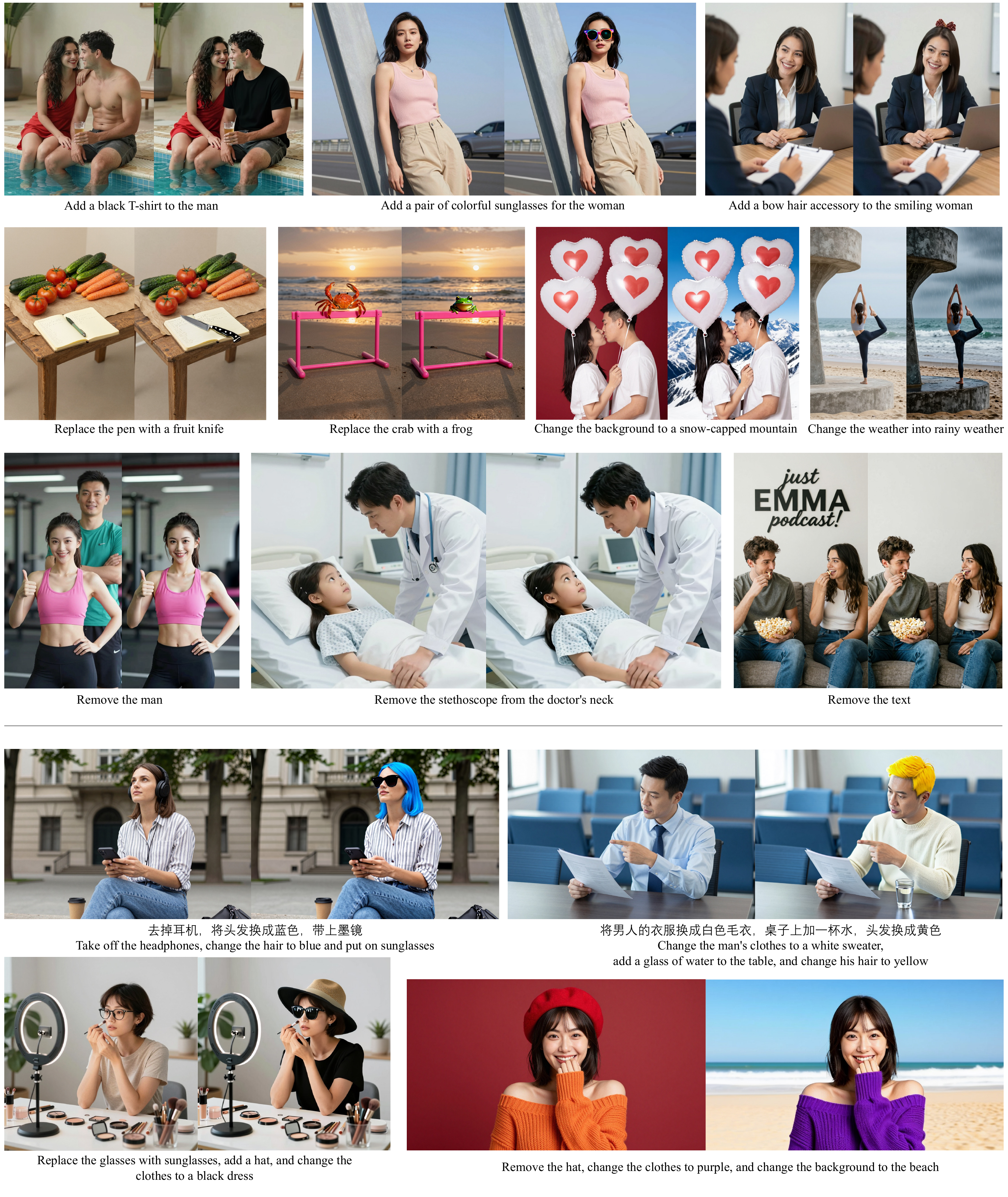}
    \caption{\textbf{Showcase of EMMA's image editing capability.} EMMA effectively adheres to editing instructions while maintaining the character consistency. Furthermore, it natively supports both Chinese editing instructions and complex instructions. It is worth emphasizing that almost all editing training data used for EMMA consisted solely of single English instructions. The above results indicate that EMMA shows strong generalization capabilities, enabling it to handle complex multimodal tasks effectively.
} 
    \label{fig:it2i_vis}
\end{figure}

During the evaluation of EMMA, we have observed several interesting phenomena: 1) without incorporating Chinese T2I generation and editing data during training, {EMMA has the ability to directly support T2I generation and editing based on Chinese instructions, as shown in Figure~\ref{fig:it2i_vis}.} This capability is likely attributed to the inclusion of Chinese data in multimodal understanding datasets, which makes the understanding branch of EMMA capable of handling Chinese instructions. 2) while trained solely on single-instruction editing data, EMMA is still capable of performing the editing followed by complex instructions, {as illustrated in Figure~\ref{fig:it2i_vis}}. The emergence of this ability is probably due to the multimodal chain-of-thought data, which enables the unified model to comprehend complex instructions and successfully execute the corresponding editing tasks.

\section{Conclusion}
In this work, we propose EMMA, an efficient multimodal unified architecture capable of performing multimodal understanding, generation, and editing tasks. Specifically, EMMA significantly reduces the number of visual tokens through an efficiently compressed autoencoder and a channel-wise token concatenation mechanism. Additionally, it employs a mixture-of-experts strategy in the visual understanding encoder to enhance visual perception. A shared-and-decoupled unified architecture is designed to mutually benefit various multimodal tasks while meeting their distinct modeling requirements. Experimental results demonstrate that EMMA outperforms current unified architectures across multiple benchmarks in multimodal understanding, generation, and editing tasks.

\clearpage

\bibliographystyle{abbrvnat}
\bibliography{main}

\end{CJK*}
\end{document}